\documentclass[final]{cvpr}

\usepackage{times}
\usepackage{epsfig}
\usepackage{graphicx}
\usepackage{amsmath}
\usepackage{amssymb}
\newcommand\norm[1]{\left\lVert#1\right\rVert}
\newcommand{\supp}[1]{\textcolor{black}{#1}}

\usepackage[pagebackref=true,breaklinks=true,colorlinks,bookmarks=false]{hyperref}

\def\cvprPaperID{7689} 
\def\confYear{CVPR 2021}
\begin{document}

\title{Anticipating human actions by correlating past with the future with Jaccard similarity measures}

\author{Basura Fernando\\
IHPC, A*STAR, Singapore.
\and
Samitha Herath\\
Dept of Data Science \& AI, Monash University\\
}

\maketitle

\begin{abstract}
We propose a framework for early action recognition and anticipation by correlating past features with the future using three novel similarity measures called Jaccard vector similarity, Jaccard cross-correlation and Jaccard Frobenius inner product over covariances. Using these combinations of novel losses and using our framework, we obtain state-of-the-art results for early action recognition in UCF101 and JHMDB datasets by obtaining 91.7 \% and 83.5 \% accuracy respectively for an observation percentage of 20. Similarly, we obtain state-of-the-art results for Epic-Kitchen55 and Breakfast datasets for action anticipation by obtaining 20.35 and 41.8 top-1 accuracy respectively.
\end{abstract}

\section{Introduction}
Action anticipation ability of humans is an evolutionary gift that allows us to perform daily tasks effectively, efficiently and safely.
This phenomena is known as mental time travel~\cite{Suddendorf2007}.
Even if humans are good at predicting the immediate future, how this happens internally inside our brain remains a mystery.
In an era where Artificial Intelligence is growing, human action anticipation has naturally become an important problem in Computer Vision. Various forms of action prediction problems have been studied in the literature such as early action recognition~\cite{gammulle2019,Kong2014,Lan2014,Ma2016,Ryoo2011,sadegh2017,shi2018action,xu2019prediction},  anticipation~\cite{Furnari2019,Miech2019,qi2017predicting} and activity forecasting~\cite{AbuFarha2018,ke2019time,piergiovanni2020adversarial}. 
The objective of Early Action Recognition~\cite{Lan2014} (EAR) (also known as action prediction) is to classify a given video from a partial observation of the action. In this case, the observed video and the full video contain the same action and methods observe about 10\%-50\% of the full video to recognize the action.
In contrast, Action Anticipation (AA) methods~\cite{Furnari2019,Miech2019}  aim at predicting a future action $\delta_t$ seconds before the future action starts and the observed video contains an action different from the future action. 
In both cases, models observe the first part of the video and predict the ongoing action or the future action.

Humans have the natural ability to correlate past experiences with what might happen in the future. 
For example, when we see someone walking toward the door inside a corridor, we can say that person will open the door with a high confidence.
Perhaps the action of walking towards the door correlates with the future action of "opening the door" with a high probability and humans may learn these associations from a very young age. 
In this paper, we propose to solve the problem of action anticipation and early prediction by correlating past features with the future. To do this, we propose a new framework and three novel loss functions.


Our framework maximizes the correlation between observed and the future video representations. By doing so, at train time our model learns to encapsulate future action information given the observed video. 
At test time our model exploits this correlation to infer future temporal information and makes accurate action predictions. Similar ideas have been explored before in the literature.
However, what remains unclear is at what abstraction one should exploit this correlation between the future and past? We argue it is better to exploit this correlation at the higher levels of the video representation and also at class level. The second question is how to maximize this correlation at higher levels of the video representation?

In this paper, we show that commonly used techniques such as minimizing the L2 distance or maximizing the vector correlation or the cosine similarity between future and observed video representations is not ideal.
Although conceptually the vector correlation makes sense, it is not bounded. 
The cosine vector similarity seems a relevant choice for this problem.
However, when used for deep representation learning, there are limitations to cosine similarity.
We discuss these limitations in detail in section~\ref{sec.jvs}. Briefly, the cosine similarity between a vector $\mathbf{z}$ and $k\mathbf{z}$ (where $k$ is a scalar) is always 1.0 (or -1.0) irrespective of the value of $k$. 
This property of cosine similarity could potentially hurt the learned representation. 
Ideally, a vector similarity measure should take into account both the magnitude and angle between vectors and the similarity should be bounded. Inspired by Jaccard Similarity overs sets, we propose Jaccard Vector Similarity (JVS) which has good properties when learning representations by maximizing the vector similarities.

Furthermore, we show that Jaccard similarity can be extended to not only vectors, but also for matrices.
Specifically, in this paper we extend Frobenius Inner Product (FIP) to work with covariance matrices and propose a new similarity measure called Jaccard Frobenius Inner Product (JFIP). 
We extract the covariance matrix of the observed and future videos and make the observed covariance matrix contains information about the future by maximizing JFIP between them.
Unlike FIP over covariance matrices, the JFIP similarity is bounded between -1 and 1 and smooth over the space of covariance matrices. 
We also propose to exploit cross-correlations between observed and future video representations and propose a new similarity measure based on Jaccard similarity over cross-correlations called Jaccard cross-correlation (JCC).
By exploiting bounded similarity measures such as JVS, JFIP and JCC, we correlate past with the future for action anticipation and early recognition using a common framework. We use slightly different architectures for EAR and AA problems. We show the benefit of Jaccard similarity measures to learn video representations suitable for future prediction tasks in an end-to-end manner.
In a summary, our contributions are as follows:

(1) We propose a common framework for action anticipation and early action recognition by exploiting the correlations between observed and future video representations. We show some limitations of cosine similarity when used for deep representation learning and propose a novel similarity measure called Jaccard Vector Similarity. We experimentally show that JVS is better than cosine similarity, vector correlation, and L2 loss for the task of correlating past features with the future for action anticipation and early action recognition.
(2) We further extend the Jaccard Similarity for covariance matrices and cross-correlation. 
We propose two novel similarity measures called Jaccard-cross-correlation and Jaccard Frobenius Inner Product over covariance matrices which performs better than standard cross-correlation, Frobenius inner product, Frobenius norm over covariance matrices, and Bregman divergence. We show the impact of these novel loss functions for action anticipation and early prediction on four standard datasets.

\section{Related work}
In early action recognition the objective is to classify ongoing action from partially observed action video.
These methods observe 10\% - 50\% of the video and then try to recognize the action ~\cite{Ryoo2011,Kong2014,sadegh2017}. These methods assume the input is a well trimmed video containing a single human action. 
Early action recognition models can be classified into various groups.
There are methods that aim to learn video representations suitable for early action recognition by handling uncertainty using new loss functions~\cite{sadegh2017,Jain2016,Ma2016}.
Second group of methods generate features for the future and then use classifiers to predict actions using generated features~\cite{shi2018action,Vondrick2016}. 
Third group of methods generate future images (either RGB or motion images) and then classify them into human actions using convolution neural networks ~\cite{Zeng2017a,Wang2017,Rodriguez2018}. 
However, RGB image generation for the future is a very challenging task, especially for a diverse video sequence.
Similarly, some methods aim to generate future motion images and then try to predict  action for the future~\cite{Rodriguez2018}. Perhaps these methods may not be able to generate details of the scenes and actions and therefore not an ideal solution for action anticipation and early action recognition.

Our method is somewhat similar to those methods that generate future features. 
However, we do not explicitly generate future features, rather we correlated future video representations with the observed data so that we can encapsulate enough information about the future in our observed video representation. Besides, we train our models end-to-end to take advantage of novel loss functions.
Somewhat similar idea to ours is the work of ~\cite{tran2019back} where they train two separate action recognition and anticipation models. They distill information from the recognition model to the anticipation model using unlabeled data. In contrast, we do not have two dedicated models for recognition and anticipation and instead of model distillation, we directly make use of future video representations to correlated the past observations with the future using a novel set of Jaccard similarity losses over vectors, cross-correlations and covariance matrices of past and the future features.

Authors in~\cite{Miech2019}, use prediction and transitional models for action anticipation. We also follow a similar idea for action anticipation. However, our model makes an explicit connection to the future features by exploiting correlations between observed and future video representations making both prediction and transitional models more effective. Authors in~\cite{Wang2020} use a transformer like architecture to encode observed video and then use progressive feature generation models to generate future features and then classify them. Interestingly, authors make use of L2 loss to minimize feature reconstruction error. Indeed, encoder-decoder type of architecture is a natural choice for action anticipation and shown to be very effective for action anticipation~\cite{Furnari2019,Gao2017}. 
Recently, an effective way to aggregate temporal information using so-called recent and spanning features for action anticipation is demonstrated in~\cite{Sener2020}. We believe the findings of \cite{Sener2020} is orthogonal to our work.
Some other methods forecast more than one action into the future after observing around 10-50\% of a long video~\cite{Mehrasa2019,AbuFarha2018,ke2019time,piergiovanni2020adversarial,Zhao2020,Ng2020}. In this paper, we only focus on short term action anticipation.

\section{Method}
\subsection{Problem statement: early action recognition}
The objective of early action recognition is to predict \emph{an ongoing action} as early as possible. 
Typically, methods observe $p\%$ of the action and then predict the category of ongoing action.
Let us define the \textbf{observed video} sequence which corresponds to $p\%$ of the action by $V_o = \left< I_1, \cdots I_t \right>$ and the video sequence which corresponds to full action by $V = \left< I_1, \cdots, I_t, \cdots, I_T \right>$ where $I_j$ is the $j^{\text{th}}$ frame of the video containing action label $y$. The objective is to predict the action $y$ of the video $V$ only by processing the first part of this video denoted by $V_o$, i.e. \textbf{observed video}.
\subsection{Our architecture for early action prediction.}
\begin{figure}[t]
\begin{center}
\includegraphics[width=0.7\linewidth]{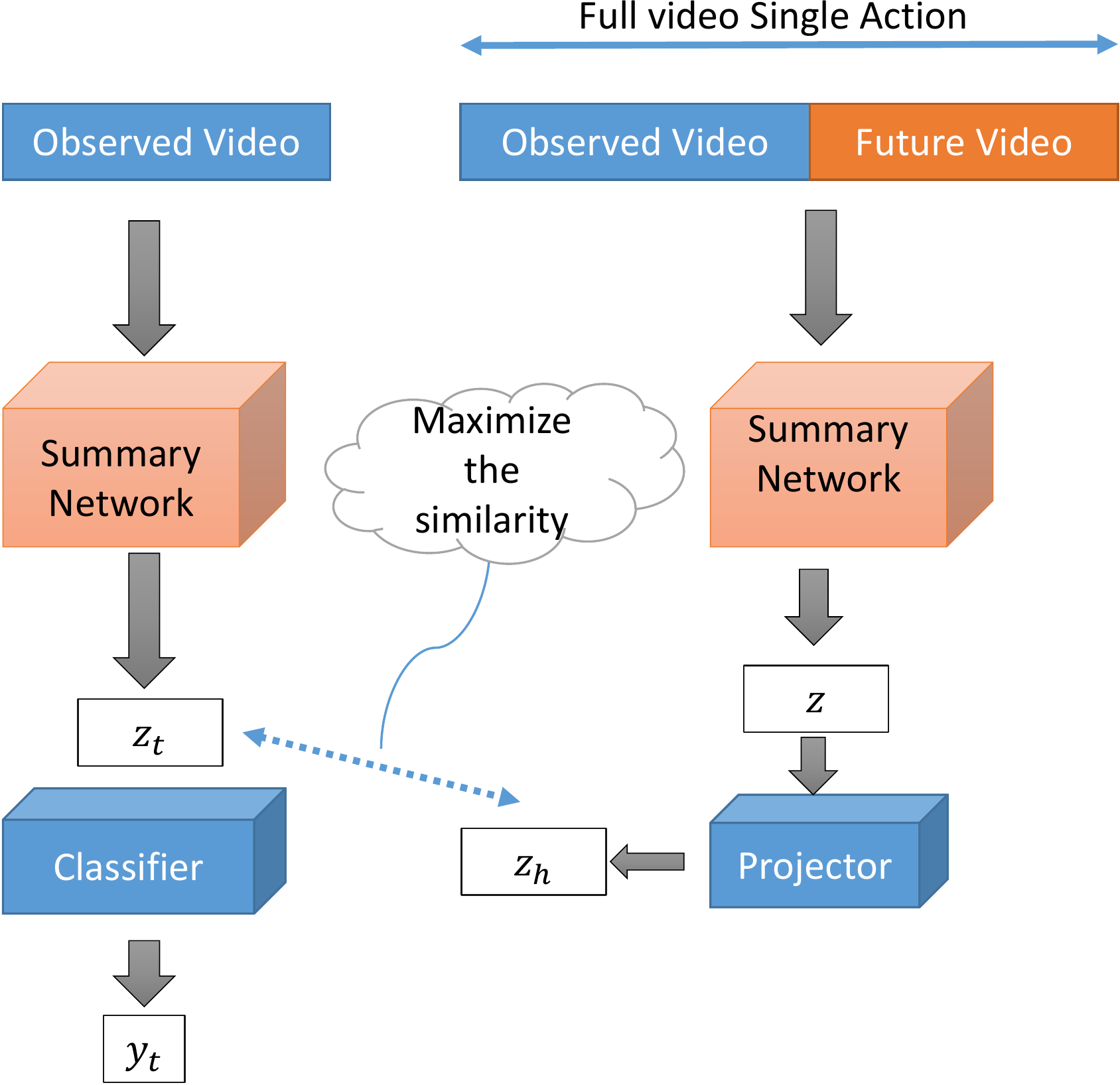}
\end{center}
\caption{High level visual illustration of early action prediction architecture.}
\label{fig:early.archi}
\end{figure}
A visual illustration of our early action prediction model is shown in Figure~\ref{fig:early.archi}.
We extract visual features from the observed video $V_o$ using an end-to-end trained \emph{feature summarizing network} (e.g. Resnet(2D+1D) or Resnet50+GRU) to get feature vector $\mathbf{z_t}$. 
The $\mathbf{z_t}$ feat summarizes the spatial temporal information of the observed video $V_o$. Similarly, we extract the feat $\mathbf{z}$ from the entire video $V$ using the same summarizing network. 

At test time we classify the feature vector from the observed video (i.e. $\mathbf{z_t}$) to get the action label $\hat{y}$. Because the observed video $V_o$ does not contain all information about the full action, during training we transfer information from the full video $V$ to the observed feature vector $\mathbf{z_t}$. 
By doing so, we aim to generate a video representation $\mathbf{z_t}$ that entails information about the future unseen video.
To do that we train our model by maximizing the similarity between $\mathbf{z_t}$ and $\mathbf{z}$. However, directly maximizing the similarity is not effective as obviously the  $\mathbf{z_t}$  and $\mathbf{z}$ are obtained from different sources of information. In-fact, $\mathbf{z}$ from the full video contains more information and our objective is to transfer maximum amount of information  to the observed feature vector $\mathbf{z_t}$ from $\mathbf{z}$. To do that we use a linear projection on $\mathbf{z}$ to obtain $\mathbf{z_h}$ and then maximize the similarity between $\mathbf{z_h}$ and $\mathbf{z_t}$. We experimentally validate different forms of non-linear functions to map $\mathbf{z} \rightarrow \mathbf{z_h}$ and found that linear mapping is the best.

We maximize the similarity between $\mathbf{z_t}$ and $\mathbf{z_h}$  by function $\phi(\mathbf{z_h}, \mathbf{z_t})$ that measures some notion of similarity between $\mathbf{z_h}$ and $\mathbf{z_t}$. Therefore, during training, we minimize the following objective function.
\begin{equation}
    L_{CE}(y,\hat{y}) + \lambda exp(-\phi(\mathbf{z_h}, \mathbf{z_t}))
    \label{eq.loss}
\end{equation}
Here $L_{CE}(y,\hat{y})$ is the cross-entropy loss and $\lambda $ is a scalar hyper-parameter. 

\subsection{Suitable loss functions}
It is important to select suitable loss function for $\phi(\mathbf{z_h}, \mathbf{z_t})$. Typically, natural choices would be to use cosine similarity or simply the vector correlation. However, we argue that a combination of vector similarity, cross-correlation and covariance measure are more suitable as these measures provide a comprehensive way of maximizing similarity between  $\mathbf{z_h}$ and $\mathbf{z_t}$. Next, we present novel similarity measures which are more suitable for the tasks.

\subsubsection{Jaccard Vector Similarity Loss}
\label{sec.jvs}
We argue that measures such as cosine similarity between vectors are not ideal when we want to learn representation by maximizing the similarity between pairs of data points. 
For example, the cosine similarity between a vector $\mathbf{z}$ and $k\mathbf{z}$ would be $1$ even if the scalar $k$ is infinitely large.
This has an implication on the learning process as two vectors that have very different magnitudes are similar with respect to the cosine similarity due to small angle.
On the other-hand measures such as L2 distance are unbounded and harder to optimize and usually do not generalize to the testing data.
To overcome this limitation of cosine-similarity, we propose a novel vector similarity measure called \textbf{Jaccard Vector Similarity}.
Typically, the Jaccard similarity is only defined over sets by computing the fraction of the cardinality of intersection set over the cardinality of the union of sets. 
We extend this concept somewhat analogically over vectors and define the Jaccard Vector Similarity as follows.
\begin{equation}
   \phi(\mathbf{z_h}, \mathbf{z_t}) = \frac{2 \mathbf{z_h} \cdot \mathbf{z_t}}{\mathbf{z_h} \cdot \mathbf{z_h} + \mathbf{z_t} \cdot \mathbf{z_t}}
\end{equation}

\begin{figure}[h!]
\centering
\includegraphics[width=0.49\linewidth]{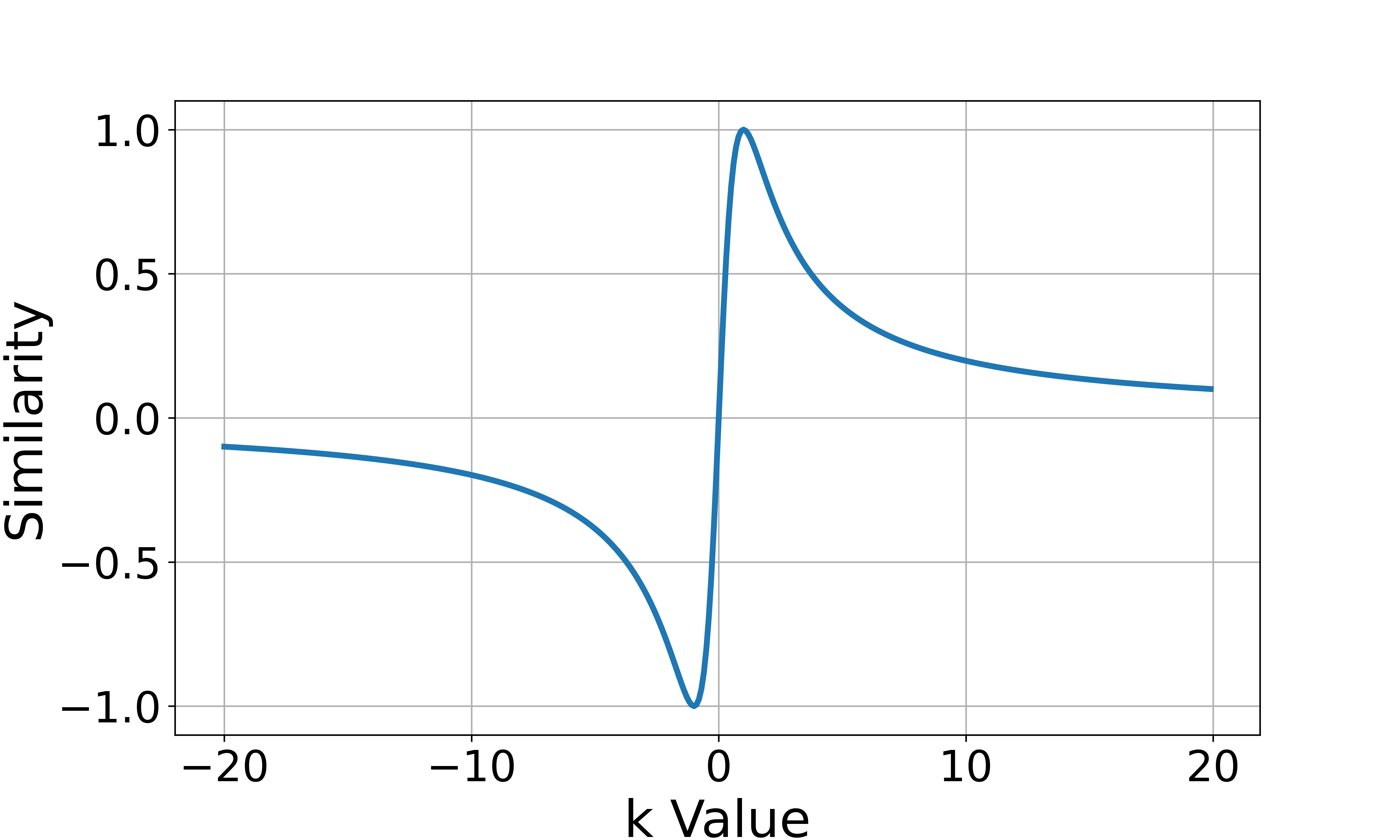}
\includegraphics[width=0.49\linewidth]{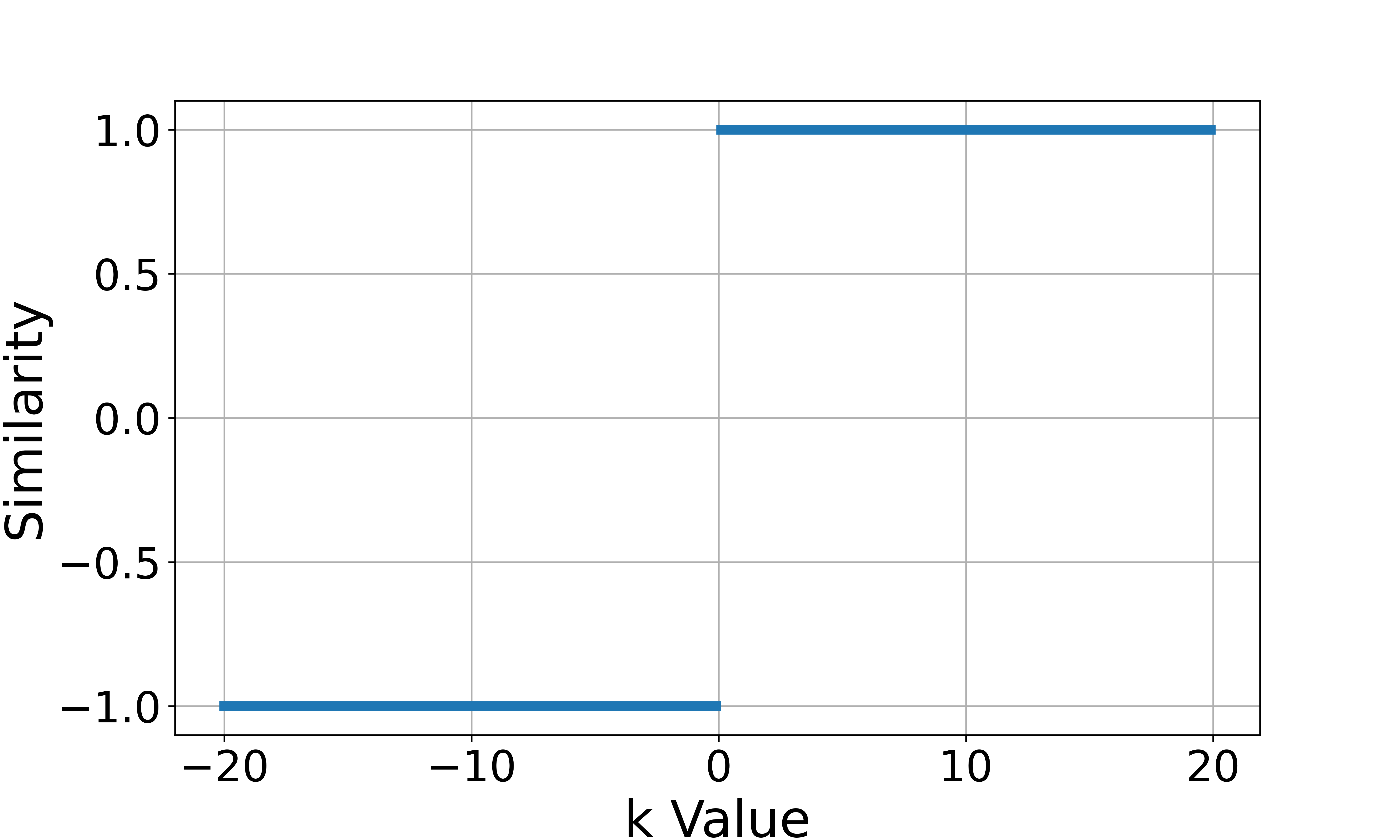}
\caption{The behavior of Jaccard Vector Similarity (left) vs Cosine Similarity (right) for two vectors $\mathbf{z}$ and $k\mathbf{z}$ where $k$ is a scalar. The cosine similarity between $\mathbf{z}$ and $k\mathbf{z}$ will be +1.0 or -1.0 irrespective the value of $k$ and cosine similarity is not smooth.}
\label{fig.prop}
\end{figure}

For a given pair of vectors $\mathbf{z}$ and  $k \mathbf{z}$ (where $k$ is a scalar), we observe the following differences in cosine similarity and the Jaccard Vector Similarity (JVS) as summarized in Table~\ref{tbl.prop}.
Illustration of the behavior of Jaccard Vector Similarity is shown in Figure~\ref{fig.prop}.
\begin{table}[h!]
\centering
\begin{tabular}{|c|c|c|} \hline
Property & Cosine & Jaccard \\ \hline
$k \rightarrow \infty$ & 1 & 0. \\ \hline
$k \rightarrow 0$ & Not defined & 0. \\ \hline
$k \rightarrow -\infty$ & -1 & 0. \\ \hline
Functional Form & $sign(K) \times 1.$ & $\frac{2k}{k^2 + 1}$  \\ \hline
\end{tabular}
\caption{Properties of Jaccard Vector Similarity for two vectors $\mathbf{z}$ and $k\mathbf{z}$ where $k$ is a scalar.}
\label{tbl.prop}
\end{table}
In this figure, you observe the behavior of Jaccard Vector Similarity for different k values. 
Both cosine and JVS is based on the vector correlation, i.e. $\mathbf{z_h} \cdot \mathbf{z_t}$. 
However the normalization of them is different.
Now imagine $\mathbf{\delta_k}$ is a small vector and the cosine similarity between $\mathbf{z}$ and $k(\mathbf{z+\delta_k})$ will be closer to 1.0 irrespective of the values of scalar $k$, where ($k>0.)$. We believe this behavior of cosine similarity is not ideal for learning representations in the context of deep learning, especially because it is not a smooth function and similarity does not depend on the magnitude. 
In contrast, Jaccard Vector Similarity considers both angle and the magnitude of two vectors to determine the bounded similarity.
Especially, the JVS is fully differentiable and a smooth function over the entire vector space.

We argue that JVS is a better option than cosine similarity, especially, for deep representation learning methods and we propose to use it for training our action prediction models where the term $\phi(\mathbf{z_h}, \mathbf{z_t})$ in Equation~\ref{eq.loss} is obtained by JVS.
Therefore, we maximize the similarity between feature vectors derived from the observed and future videos using JVS.
This allows us to transfer information from full video to the observed video during training. In other words, we correlate future information with past observations to make effective future predictions.

\subsubsection{Jaccard Cross-Correlation Loss.}

Furthermore, we extend the similarity maximization to cross-correlations and covariances. In this case, the input to similarity function $\phi(\cdot, \cdot)$ is either a cross-correlation matrix or a covariance matrix obtained from the batch data. 
For a given batch of $n$ videos, we obtain observed data matrix $Z_t$ and the full feature matrix $Z$ where each row in these matrices are obtained from the corresponding feature vectors ($\mathbf{z_t}$, $\mathbf{z}$) of each video in the batch.
As before, we project the full video features $\mathbf{z}$ to $\mathbf{z_h}$ by linear mapping and construct matrix $Z_h$.

Then the cross-correlation matrix of $Z_h$ and $Z_t$ is obtained by $E[Z_h^T \times Z_t]$ \cite{gubner2006probability} where $E[]$ is the expectation. Then we define the similarity loss based on the cross-correlation matrices and the Jaccard similarity as follows:
\begin{equation}
  \norm{exp(- 2 \frac{E[Z_h^T \times Z_t]}{E[Z_h^T \times Z_h] +E[ Z_t^T  \times Z_t]})}_{mean}
\end{equation}
where $|| C ||_{mean}$ is the mean norm of the matrix define by $\frac{1}{n^2} \sum_{i=1}^{n} \sum_{j=1}^{n} C_{i,j}$. We call this measure as the \textbf{Jaccard Cross-Correlation} (JCC) loss.
The objective of this JCC loss is to maximize the cross-correlation between observed and future data and therefore to transfer more (future) information to the observed video representation.

\subsubsection{JFIP: Jaccard Frobenius Inner Product Loss.}

Finally, we also propose a new similarity measure by extending the Frobenius inner product using Jaccard similarity.
For this we use covariance information of observed and future videos.
Let us define the covariance matrix of $Z_h$ by $C_h$ and the covariance matrix of $Z_t$ is $C_t$.
The new \textbf{Jaccard Frobenius Inner Product} of covariance matrices is defined as follows:
\begin{equation}
    \phi(Z_h, Z_t) = \frac{2\left< C_h,C_t \right>_F}{\left< C_h,C_h \right>_F + \left< C_t,C_t\right>_F}
\end{equation}
where $\left<C_a, C_b\right>_F$ denotes the Frobenius inner product between matrices $C_a$ and $C_b$.
Similar to Jaccard vector similarity, the Jaccard Frobenius Inner Product (JFIP) has bounded similarity and
has a nice smoothness property which is useful for deep representation learning tasks.
Specifically, the main idea of JFIP is to make sure that the observed video representation contains information about the second order statistics of the future video data. In other words, we aim to transfer covariance information of the future to the past representation.

We propose to make use of Jaccard vector similarity (JVS), Jaccard cross-correlation (JCC) and Jaccard Frobenius inner product (JFIP) for optimizing the representation using the loss function in equation~\ref{eq.loss}.
While JVS provides a direct measure of similarity between observed and full videos, JCC and JFIP losses use higher order statistical information between  observed and full video features.
All these losses provide complementary information and we argue that "Jaccard" measures are bounded and smooth and therefore effective in learning representation when we maximize similarity. Therefore these are better than traditional measures such as vector correlation, cosine similarity, L2 distance, traditional cross-correlation, Frobenius inner product and Bregman divergence~\cite{Harandi2014} over covariance matrices. Once we define the similarity function, to transform them to a loss function, we use the negative exponential $exp(-\phi())$ function as shown in Table~\ref{tbl.losses} when needed.

\subsection{Action anticipation with Jaccard losses.}
In this section we further extend our architecture for action anticipation. In action anticipation the objective is to predict an action $\delta_t$ seconds before action starts.
Let us assume that the length of the observed video is $t_o$ seconds and then the next action starts at $t_o+\delta_t$ seconds.
As before, let us denote the observed video by $V_o$ and the future action video by $V_f$.
Let us also define the label of observe action by $y_o$ and the label of the future action by $y_f$. 
We predict the next future action $y_f$ by processing the observed video $V_o$ which ends $\delta_t$ seconds before the action $y_f$ starts.

\begin{figure}[t]
\begin{center}
\includegraphics[width=0.6\linewidth]{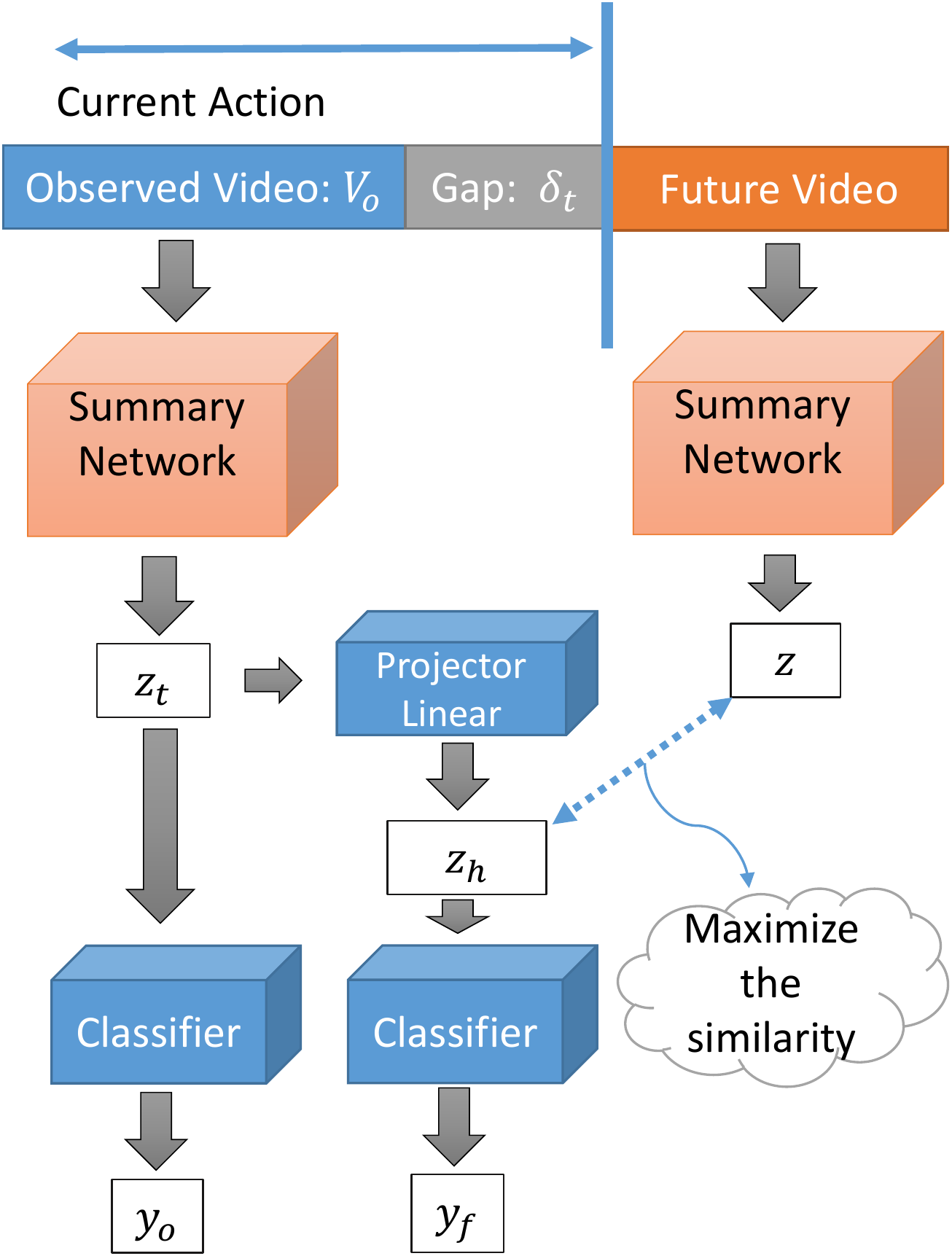}
\end{center}
\caption{High level visual illustration of action anticipation architecture.}
\label{fig:anti.archi}
\end{figure}

\subsection{Architecture for action anticipation.}
A high level illustration of our action anticipation architecture is shown in Figure~\ref{fig:anti.archi}.
We obtain the feature representation $\mathbf{z_t}$ from the observed video and then classify it to obtain the observe action $\hat{y_o}$ using a linear softmax classifier. 
We transform $\mathbf{z_t}$ to $\mathbf{z_h}$ using a linear projection. 
The vector $\mathbf{z_h}$ essentially simulates the feature representation for the future action $y_f$.
In-fact, we classify $\mathbf{z_h}$ using the same action classifier as before to obtain $\hat{y_f}$. 

During training, we also extract the \emph{actual} feature representation of the future video denoted by $\mathbf{z}$ and maximize the similarity between $\mathbf{z}$ and $\mathbf{z_h}$ using JVS, JCC and JFIP as before. 
If we denote the video summary network by $g()$, then $\mathbf{z} = g(V_f)$ and $\mathbf{z_t} = g(V_o)$. 
Typically we limit the length of $V_o$ and $V_f$ to be short clips of 2 seconds. The anticipation gap ($\delta_t$) is usually set to 1 second.
Now loss function consists of two cross-entropy losses as follows:
\begin{equation}
    L_{CE}(y_o,\hat{y_o}) + L_{CE}(y_f,\hat{y_f}) + \lambda exp(-\phi(\mathbf{z_h}, \mathbf{z}))
    \label{eq.loss.anti}
\end{equation}
where $L_{CE}(y_o,\hat{y_o})$ is the cross-entropy loss for observed action and $L_{CE}(y_f,\hat{y_f})$ is the cross entropy loss for the future action label.
The term $exp(-\phi(\mathbf{z_h}, \mathbf{z}))$ is obtained by JVS, JCC and JFIP.

It is intuitive to assume that there is a significant bias when predicting next action if we know the previous action.
For example, if we observe action "open door", it is more likely to see an future action such as "turn on lights".
To explore this bias in action transition space, we propose to make use of a linear projection that allows us to predict the next action from the observed action.
Therefore, we also predict the next action (denoted $\hat{y_{of}}$) by linearly projecting the observed action score vector.
Therefore, during training we have the following multi-task training loss.
\begin{equation}
    L_{CE}(y_o,\hat{y_o}) + L_{CE}(y_f,\hat{y_{of}}) + L_{CE}(y_f,\hat{y_f}) +\lambda exp(-\phi(\mathbf{z_h}, \mathbf{z}))
    \label{eq.loss.anti}
\end{equation}
We optimize the above objective function for action anticipation.
The loss term $L_{CE}(y_o,\hat{y_o})$ makes sure that the classifier in Figure~\ref{fig:anti.archi} is well trained for observed actions.
The term $L_{CE}(y_f,\hat{y_{of}})$ makes sure that the future action prediction obtained by linearly transforming the observed action is accurate.
The term $\lambda exp(-\phi(\mathbf{z_h}, \mathbf{z})$ makes sure that $\mathbf{z_h}$ is highly correlated with the actual future action representation $\mathbf{z}$. 
This correlation helps to predict both $\hat{y_{of}}$ and $\hat{y_{f}}$ accurately.
At test time, we take the sum of scores $\mathbf{\hat{y_{of}}}$ and $\mathbf{\hat{y_f}}$ to get the correct future action prediction score.

\subsection{Alternative loss functions}

Now we present some alternative loss functions that we evaluate in our experiments to demonstrate the effectiveness of JVS, JCC  and JFIP loss.
For this discussion let us assume that observed video representation is $\mathbf{z_t}$ and the future representation is $\mathbf{z}$. The objective of these loss functions is to minimize the differences between $\mathbf{z_t}$ and $\mathbf{z}$ or to maximize the similarity.
For batch wise measures such as covariance and cross-correlation measures, we denote the observed batch of video representation by matrix $Z_t$ where each row represents a vector from a video.
Similarly, $Z$ is the representation of the future video.
The covariance matrix of the observed batch obtained from $Z_t$ is denoted by $C_t$ and the covariance matrix obtained from $Z$ is denoted by $C_z$.
In Table~\ref{tbl.losses} we present all loss functions we compare in our experiments.
\begin{table}[t!]
    \centering
    \resizebox{\columnwidth}{!}{%
    \begin{tabular}{l|l} \hline
         Name & Loss Equation \\ \hline
         Correlation (Corr.) & $exp(- \mathbf{z_t} \cdot \mathbf{z} )$ \\
         Cosine & $exp(- \frac{\mathbf{z_t} \cdot \mathbf{z}}{ ||\mathbf{z_t}||  \times ||\mathbf{z}||  })$ \\
         L2  & $exp(||\mathbf{z_t - z}||)$ \\
         Jaccard vector similarity (JVS) & $exp(- 2\frac{\mathbf{z_t} \cdot \mathbf{z}}{ \mathbf{z_t}\cdot\mathbf{z_t}  + \mathbf{z} \cdot \mathbf{z}  })$ \\ \hline
         \multicolumn{2}{c}{Cross-correlation measures} \\\hline
         Cross-correlation (CC) &  $||exp(- E[Z_t^T \times Z])||_{mean}$ \\
         JCC &  $\norm{exp(- 2 \frac{E[Z^T \times Z_t]}{E[Z^T \times Z ]+ E[Z_t^T  \times Z_t]})}_{mean}$ \\\hline
         \multicolumn{2}{c}{Covariance measures} \\\hline
         Frobenius norm (FN) & $exp(|| C_t - C_z  ||_F)$ \\
         Bregman divergence (BD)~\cite{Harandi2014} & $Trace\{C_t C_{z}^{-1} + C_{z}C_{t}^{-1}\}-d$ \\
         Frobenius inner product (FIP) & $exp(- \left<C_z, C_t\right>_F  ) $ \\
         JFIP & $exp( -\frac{2\left< C_z,C_t \right>_F}{\left< C_z,C_z \right>_F + \left< C_t,C_t\right>_F})$\\ \hline
    \end{tabular}
    }
    \caption{Different loss functions used in this paper.}
    \label{tbl.losses}
\end{table}

\section{Experiments}
In this section we evaluate our early action prediction and action anticipation models. First, we evaluate the impact of novel loss functions for early action anticipation in section~\ref{sec.exp.early} and then present action anticipation results in section~\ref{sec.exp.anticipation}.

\subsection{Early action prediction results}
\label{sec.exp.early}
We use UCF101~\cite{Soomro2012} and JHMDB~\cite{Jhuang2013} datasets to evaluate early action prediction. First, we evaluate the impact of loss functions presented in Table~\ref{tbl.losses}.
We use Resnet50 as the frame level feature extractor and then use a GRU model to summarize the temporal information of the observed and future features to implement the architecture presented in Figure~\ref{fig:early.archi}.
The last hidden state of GRU would generate $\mathbf{z_t}$ and $\mathbf{z}$.
We use Adam optimizer with a learning rate of 0.001 and batch size of 64 videos. Resnet50 model is pretrained on the ImageNet dataset for image classification.
We use both RGB and optical flow frames as the input in a two-stream architecture for JHMDB dataset.
However, we use only the RGB stream for the UCF101 dataset to complete all ablation studies.
We set $\lambda$ to be $1.0$ for all bounded losses and $0.001$ unbounded losses (such as for Frobenius norm loss, L2 loss, Cross-correlation loss, Bregman divergence and Frobenius inner product). Setting $\lambda$ to larger values for unbounded losses causes the network to stop learning. Experimentally, we evaluate a range of values for $\lambda$ and found that $0.001$ is good for unbounded losses.

In early action prediction, we observe 10\% or 20\% of the video and then predict the action of the video.
Some UCF101 videos are very long and therefore following prior work, if the video is longer than 250 frames, we only observe a maximum of 50 frames.
Our baseline model simply classifies the video using observed video representation $\mathbf{z_t}$ using a linear classifier.
This is equivalent to setting $\lambda=0$ in equation~\ref{eq.loss}.
We report results for various loss functions in Table~\ref{tbl.early.results}.
\begin{table}[t!]
\footnotesize
\centering
\begin{tabular}{|l|c|c| c|}\hline
 dataset & \multicolumn{2}{c}{JHMDB} & UCF101 (RGB) \\ \hline
 observation (\%) &  10\% & 20\% &  20\%\\ \hline
 Baseline		& 49.0 & 50.4 & 66.4\\ \hline
 \multicolumn{4}{|c|}{Vector similarity measures} \\ \hline
 Corr.			& 47.8 & 51.3 & 66.4 \\ 
 Cosine         & 51.8 & 54.5 & 66.9\\
 L2             & 50.4 & 51.5 & 67.5 \\ 
 JVS            & \textbf{62.6} & \underline{64.7} & \textbf{72.3}\\  \hline
 \multicolumn{4}{|c|}{Cross-correlation measures} \\ \hline
Cross-correlation & 49.2  & 49.8 & 66.0\\
JCC               & \underline{60.3}  & \textbf{66.1} & \underline{70.2} \\ \hline
\multicolumn{4}{|c|}{Covariance measures} \\ \hline
Frobinious norm     & 53.3  & 55.0 & 69.9\\
Bregman divergence  & 51.1  & 44.1 & 64.4 \\
Frobenius inner product  & 53.4 & 60.5 & 66.2 \\
JFIP & \underline{56.4} & \underline{64.4} & \underline{71.4}\\ \hline
\end{tabular}
    \caption{Results using Resnext50 as the base model.}
    \label{tbl.early.results}
\end{table} 

As we can see from the results, our Jaccard Vector Similarity (JVS) loss obtains better results than correlation, cosine similarity and L2 losses. 
Cosine similarity and L2 losses perform better than the baseline model indicating that our architecture of early action anticipation is useful even under these losses.
Interestingly, JVS loss is 10.8\% better than Cosine loss on JHMDB dataset for observation percentage of 10\% indicating the impact of this new bounded and fully differentiable loss. Similar trends can be seen for UCF101.
The main advantage of JVS is that it takes into account both the angle and the magnitudes of two vectors into consideration while keeping good properties of cosine similarity and makes sure the loss is differentiable at all points in the space.
These properties are very important when we learn representations by comparing similarity between two vectors using deep learning.
JVS reports a massive improvement over the baseline model.

Similarly, when cross-correlation is used, the JCC loss outperforms the classical cross-correlation. The best results for 20\% observation on JHMDB dataset is obtained by JCC loss (66.1\%).
The best results for covariance matrices-based losses is obtained by JFIP outperforming Frobenius norm loss, the Frobenius inner product loss and the Bregman divergence.

Next we evaluate the complementary nature of each of these losses by combining these losses in a multi-task manner. In Table~\ref{tbl.early.results.fusion}, we combine all Jaccard losses and then compare with traditional measures.
\begin{table}[t!]
\centering
\begin{tabular}{|l|c|c|c|}\hline
dataset & \multicolumn{2}{c}{JHMDB} & UCF101 \\ \hline
Observation (\%) &  10\% & 20\% & 20\%  \\ \hline
Baseline                &  49.0 & 50.4 &  78.9\\
FN + L2 + Cosine       &  57.8 & 64.9 &  82.0 \\
JVS + JCC + JFIP       &  \textbf{64.3} & \textbf{69.1} & \textbf{85.8}\\
\hline
\end{tabular}
    \caption{Combined losses using rgb+optical flow with Resnet50 as the base model on JHMDB and UCF101 datasets.}
    \label{tbl.early.results.fusion}
\end{table}  
We see that combination of Frobenius norm, L2 and cosine losses (FN + L2 + Cosine) performs much better than individual loss performances. 
This observation can be seen in both datasets.
These individual losses provide different properties into the learned representation and it is interesting to see that they are complementary.
For this comparison we do not include the Cross-correlation loss as it's performance is poor.
Interestingly, combination of Jaccard (JVS + JCC + JFIP) losses improves results of individual models by a significant margin and it outperforms (FN + L2 + Cosine) combination of losses.

Now we compare our results with state-of-the-art early action prediction models using effective Resnet18(2D+1D) network where it is pre-trained on Kinetics~\cite{kay2017kinetics} dataset for action classification. We use both optical flow and rgb stream to train our models end-to-end. Results are reported in Table~\ref{tbl.early.results.soa.jhmdb} for JHMDB and Table~\ref{tbl.early.results.soa.ucf} for UCF101.

\begin{table}[t!]
    \centering
    \begin{tabular}{|l|c|}\hline
Method &   Accuracy  \\ \hline
Baseline      & 61.9 \\
MM-LSTM~\cite{sadegh2017}      & 55.0 \\  
PCGAN~\cite{xu2019prediction}  & 67.4 \\
RBF-RNN~\cite{shi2018action}   & 73.0 \\
I3DKD~\cite{tran2019back}      & 75.9 \\
RGN-Res-KF~\cite{zhao2019}     & 78.0 \\ \hline    
JVS + JCC + JFIP               & \underline{82.0} \\
FN + L2 + Cosine              & 76.9 \\
JVS + JCC + JFIP+ FN + L2 + Cosine      & \textbf{83.5} \\
\hline
\end{tabular}
    \caption{Combined models using rgb+optical flow, Resnet18(2D+1D) as the base model. Comparing our results with state-of-the-art methods in the literature for observation percentage of 20\% on JHMDB dataset.}
    \label{tbl.early.results.soa.jhmdb}
\end{table}  

\begin{table}[t!]
    \centering
    \begin{tabular}{|l|c|}\hline
Method &   Accuracy  \\ \hline
Baseline (Resnet50)            & 78.9 \\
Baseline (Resnet18(2D+1D))     & 87.1 \\
MM-LSTM~\cite{sadegh2017}      & 80.3 \\     
AA-GAN~\cite{gammulle2019}     & 84.2 \\
RGN-KF~\cite{zhao2019}         & 85.2 \\ \hline   
JVS + JCC + JFIP (Resnet50)    & \textbf{85.8} \\
JVS + JCC + JFIP  (Resnet18(2D+1D))   & \textbf{91.7} \\
\hline
\end{tabular}
    \caption{Combined model using rgb+optical flow, Resnet18(2D+1D) as the base model. Comparing our results with state-of-the-art methods in the literature for observation percentage of 20\% on UCF101 dataset.}
    \label{tbl.early.results.soa.ucf}
\end{table}

On JHMDB dataset our method obtains significant improvement over prior state-of-the-art methods, specially over the recent methods (e.g. RGN-Res-KF~\cite{zhao2019} and I3DKD~\cite{tran2019back}).
Our (JVS + JCC + JFIP) model improves baseline results from 61.9\% to 82.0\% while  (FN + L2 + Cosine) obtains only 76.9\%.
Interestingly, combination of all six losses obtains 83.5\%.
Similar trend can be seen for the UCF101 dataset. Our Resnet50 baseline obtains 78.9\% while our (JVS + JCC + JFIP) model obtains 85.8 outperforming prior state-of-the-art results even using Resnet50 model. When (Resnet18(2D+1D)) is used, we obtain 91.7\% accuracy which also improves the baseline results by 4.6\%.
All these results can be attributed to the impact of novel JVS, JCC and JFIP losses and the effect of end-to-end training with our early action prediction architecture.
\subsection{Action anticipation results}
\label{sec.exp.anticipation}
We evaluate the action anticipation model using Breakfast~\cite{Kuehne2014} and Epic-Kitchen55~\cite{Damen2018} datasets.
We follow the protocol of~\cite{Miech2019,AbuFarha2018} for action anticipation using all splits on Breakfast dataset.
Similarly, we use the validation set proposed in~\cite{Furnari2019} to evaluate the performance on Epic-Kitchen55 dataset.
We use an observation window of two seconds (i.e. the length of $V_o$) and a gap of one second (i.e. the length of $\delta_t$) by default (see Figure~\ref{fig:anti.archi}) for both datasets.
We only use the Resnet18(2D+1D) model as the video summarizing network where we use the same training parameters as in early action prediction experiments. 
We train action, noun and verb predictors separately for Epic-55 dataset. Then we obtain action predictions from verb and noun predictors as well. These are averaged with the action predictor to get the final action anticipation result. 
\supp{For more information please see the supplementary material section 1.}
Following prior work~\cite{Sener2020}, we also evaluate using dense trajectories and Fisher Vectors (FV) only for Breakfast dataset.
\supp{Details of the FV architecture is given in supplementary material section 2.}
Now we demonstrate the effect of our action anticipation architecture using Epic-Kitchen and Breakfast datasets in Table~\ref{tbl.anticipation.results} where we compare our novel loss functions against standard loss measures.
\begin{table}[t!]
\centering
\begin{tabular}{|l|c|c|c|c|}\hline
 Dataset & \multicolumn{2}{|c|}{Breakfast} & \multicolumn{2}{|c|}{Epic-Kitchen} \\
 \hline
 Modality & FV & (R(2D+1D)) & \multicolumn{2}{c|}{(R(2D+1D))} \\ \hline
 Measure & \multicolumn{2}{c|}{Accuracy} &  Top 1 & Top 5\\ \hline
Baseline		& 23.4 & 24.3 & 9.82 &  24.48 \\ 
Eq.\ref{eq.loss.anti} ($\lambda=0.0$) & 23.9 & 24.6 & 10.01 & 24.82 \\ \hline
 \multicolumn{5}{|c|}{Vector similarity measures} \\ \hline
 Corr.			& 24.0 & 24.6 & 11.05   & 25.02 \\ 
 Cosine         & 24.7 & 24.8 & 10.59 &	25.56 \\
 L2             & 24.3 & 26.1 &	10.28 & 26.48\\ 
 JVS            & \underline{28.6} & \underline{28.0} & \textbf{15.20} & \underline{32.54} \\ \hline
 \multicolumn{5}{|c|}{Cross-correlation measures} \\ \hline
CC & 24.2 & 25.1 & 10.44 & 24.48\\ 
JCC               & \underline{28.6} & \underline{28.1} & \underline{14.12} & \underline{32.16}\\ \hline
\multicolumn{5}{|c|}{Covariance measures} \\ \hline
FN     & 20.4 & 21.0 & 10.21 & 23.87\\ 
BD  & 11.2 & 23.9 & 10.05 & 24.94\\ 
FIP  & 25.0 & 26.0  & 10.44 & 23.71 \\ 
JFIP & \textbf{30.3} & \textbf{30.9}   & \underline{13.89} & \textbf{33.31} \\ \hline
\end{tabular}
    \caption{Action anticipation results using our loss functions and architecture. We compare Jaccard losses (JVS, JCC, JFIP) against other loss measures for action anticipation on Breakfast and Epic-Kitchen55 datasets. (R(2D+1D)) model uses RGB stream only.}
    \label{tbl.anticipation.results}
\end{table} 

We observe that all loss functions improve over the baseline model. The baseline model directly predicts the future action from the observed video. When we set $\lambda=0.0$, the results improves only slightly.
The JVS loss performs the best out of all vector similarity measures and the improvement obtained by JVS is significant.
JFIP outperforms all other covariance measures and JCC outperforms cross-correlation loss. This trend can be seen for both Breakfast and Epic-Kitchen datasets.
The best top-1 action anticipation results are obtained by JVS loss while the best top-5 results are obtained by JFIP loss on Epic-Kitchen dataset.
Perhaps the JFIP is better at modeling the uncertainty than the JVS loss. The JFIP also obtains the overall best results in Breakfast dataset.
%
\begin{table}[t!]
    \centering
    \begin{tabular}{|l|c|c|}\hline
Method &   Top 1 & Top 5  \\ \hline
Baseline (Resnet18(2D+1D))  &   10.44 & 25.63 \\
RL~\cite{Ma2016}  & -- & 29.61 \\
VN-CE~\cite{Damen2018} & 5.79 &  17.31\\
SVM-TOP3~\cite{Furnari2018} & 11.09 & 25.42 \\
RU LSTM~\cite{Furnari2019}  & -- & 35.32 \\
FIA~\cite{zhang2020egocentric} & 14.07 & 33.37 \\
AVHN~\cite{kapidis2019multitask} & 19.29 & 35.91 \\
Our model (Resnet18(2D+1D)) & \textbf{20.35} & \textbf{39.15}  \\
\hline
\end{tabular}
    \caption{Comparing our results with state-of-the-art methods in the literature for temporal gap $\delta_t=1.0$ seconds on Epic-Kitchen55 dataset. We use rgb+optical flow, Resnet18(2D+1D) as the base model. Our model uses a combination of JVS, JCC and JFIP.}
    \label{tbl.anti.results.epic}
\end{table}
\begin{table}[t!]
    \centering
    \begin{tabular}{|l|c|}\hline
Method &   Accuracy  \\ \hline
Baseline (Resnet18(2D+1D))     & 24.3 \\
CNN of~\cite{AbuFarha2018} & 27.0\\
RNN of~\cite{AbuFarha2018} & 30.1\\
Predictive + Transitional (AR)~\cite{Miech2019} & 32.3 \\
TAB~\cite{Sener2020} (FV) & 29.7 \\
TAB~\cite{Sener2020} (I3D) & 40.1 \\ \hline
Our model (FV) & 33.6\\
Our model (Resnet18(2D+1D)) & 34.6\\
Our model (FV+Resnet18(2D+1D)) & \textbf{41.8}\\
\hline
\end{tabular}
    \caption{Comparing our results with state-of-the-art methods in the literature for temporal gap $\delta_t=1.0$ seconds on Breakfast dataset. Our model uses a combination of JVS, JCC and JFIP.}
    \label{tbl.anti.results.soa.brk}
\end{table}

We compare our model that uses combination of (JVS, JCC, and JFIP) losses with the state-of-the-art methods for Epic-Kitchen55 and Breakfast datasets in Tables~\ref{tbl.anti.results.epic} and ~\ref{tbl.anti.results.soa.brk} respectively for action anticipation task\footnote{By the time of the submission, the evaluation server on Epic-Kitchen55 test set was not available and therefore we report results only on the validation set and compare with papers that use the same protocol.}.

For Breakfast dataset, we use rgb based Resnet18(2D+1D) and Fisher vectors.
For Epic-Kitchen55 dataset, we use rgb, optical flow and object streams. 
Objects are detected using Mask-RCNN~\cite{He2017} model and spatial features are extracted using respective regions and pooled by taking the average of those features. In this case, the feature extractor is a ResNeXt-50-32x4d~\cite{Xie2017} model. Specially, this object feature stream allows us to identify nouns accurately.
All three streams are trained separately and then fused at test time by taking the average of prediction scores.

As can be seen from the results in Table~\ref{tbl.anti.results.epic}, our method obtains state-of-the-art results on Epic-Kitchen55 dataset.
Specially, it is important to note that our top-5 accuracy is considerably better than other methods.
Our model outperforms the baseline model which also uses a three stream approach (rgb+optical-flow+object) by a large margin.

We can see a similar trend in the Breakfast dataset as shown in results in Table~\ref{tbl.anti.results.soa.brk}.
As before the improvement over the baseline model is quite significant.
Our method outperforms recent methods such as~\cite{AbuFarha2018,Miech2019,Sener2020} while obtaining improvement over these prior state-of-the-art models on the Breakfast dataset.
Specifically, we do not use action segmentation methods as done in~\cite{Sener2020}. It should be noted that with action segmentation, the recent method in~\cite{Sener2020} obtains 47.0 \% on the Breakfast dataset.
However, when Fisher Vector and I3D is used without segmentation, our method outperforms ~\cite{Sener2020}.
Our method also performs better than ~\cite{Sener2020} when we use Fisher Vectors only (29.7 vs 33.6).

We have demonstrated state-of-the-art action anticipation results on two different kinds of datasets (Epic-Kitchen and Breakfast).
While Epic-Kitchen is an ego-centric action analysis dataset, Breakfast is a third person video capture dataset. As can be seen from the comprehensive set of experiments done in Tables~\ref{tbl.anticipation.results}, \ref{tbl.anti.results.epic} and \ref{tbl.anti.results.soa.brk}, our model and new loss functions are quite effective in both types of video data for action anticipation.

\noindent
\textbf{Ablation:}
Please see the supplementary material for additional experiments on the impact of loss functions and architecture.
Specifically, we evaluate the impact of L2 + Cosine similarity against the JVS loss. Furthermore, we evaluate the impact of $exp(-\phi(\mathbf{z_h}, \mathbf{z}))$ of equation \ref{eq.loss.anti} and observation branch on action anticipation.

\section{Conclusion}
We presented a framework for early action recognition and action anticipation by correlating the past with the future representations using three effective and novel loss functions.
The proposed Jaccard Vector Similarity is consistently better than vector correlation, L2 distance and cosine similarity for learning representations with our framework.
Similarly, novel Jaccard Cross-correlation and Jaccard Frobenius Inner product losses are also more effective than compared higher order losses such as Frobenius norm, Frobenius Inner product and considered Bregmen Divergence.
These Jaccard losses are smooth, fully differentiable and bounded. They are good for representation learning tasks.
These new Jaccard losses can be useful for other applications in deep learning as well, and in the future, we aim to investigate both theoretical and application aspects of these measures.
\newline
\textbf{Acknowledgment}
This research/project is supported by the National Research Foundation, Singapore under its AI Singapore Programme (AISG Award No: AISG-RP-2019-010).
{\small
\bibliographystyle{ieee_fullname}
\bibliography{egbib}
}

\end{document}